\documentclass[journal]{IEEEtran}

\usepackage{amsmath}
\usepackage{array}
\usepackage{balance}
\usepackage{color}
\usepackage{mathtools}
\usepackage{amssymb}
\usepackage{optidef}
\usepackage{amsthm}
\usepackage{mathtools}
\usepackage{setspace}
\usepackage{subfigure}
\usepackage{algorithm}
\usepackage[bottom]{footmisc}
\usepackage{tikz}
\usepackage{algpseudocode}
\usepackage{scalerel,stackengine}
\usepackage{cite}
\usepackage{booktabs}

\ifCLASSOPTIONcompsoc
 \usepackage[caption=false,font=normalsize,labelfont=sf,textfont=sf]{subfig}
\else
 \usepackage[caption=false,font=footnotesize]{subfig}
\fi

\DeclarePairedDelimiter\floor{\lfloor}{\rfloor}

\newcommand\widebar[1]{\mathop{\overline{#1}}}

\newcommand{\bfY}{\mathbf{Y}}
\newcommand{\bfR}{\mathbf{R}}
\newcommand{\bfC}{\mathbf{C}}
\newcommand{\bfx}{\mathbf{x}}
\newcommand{\bfr}{\mathbf{r}}
\newcommand{\bfE}{\mathbf{E}}
\newcommand{\bfG}{\mathbf{G}}
\newcommand{\bfZ}{\mathbf{Z}}
\newcommand{\bfs}{\mathbf{s}}

\newcommand{\bbR}{\mathbb{R}}

\newcommand{\rmC}{\mathrm{C}}

\newcommand{\rmT}{\mathrm{T}}
\newcommand{\rmR}{\mathrm{R}}
\newcommand{\rmY}{\mathrm{Y}}
\newcommand{\rmZ}{\mathrm{Z}}

\newcommand{\eff}{\mathbf{\mathcal{E}}}
\newcommand{\efff}[1]{\mathcal{E}_{#1}}
\newcommand{\scenecomp}{\mathbf{e}_\bfR(Z)}
\newcommand{\scenecomphat}{\widehat{\mathbf{e}}_\bfR(\rmZ)}

\newcommand{\tgtdis}[1]{p(\bfx_{#1}|\bfR, \rmZ)}
\newcommand{\tgtdisNoIndex}{p(\bfx|\bfR, \rmZ)}
\newcommand{\LTwoIndex}[2]{\mathcal{L}(\bfR_{#1}|\bfx_{#2},\rmZ)}
\newcommand{\LXIndex}[1]{\mathcal{L}(\bfR|\bfx_{#1},\rmZ)}
\newcommand{\LNoIndex}{\mathcal{L}(\bfR|\bfx,\rmZ)}
\newcommand{\PTransition}{\mathcal{T}_{\bfx_t}\left(\bfx_{t+1}\right)}
\newcommand{\ProposalDist}{p'(\bfx_{t+1}|\bfR,\rmZ)}
\newcommand{\Obs}{\mathcal{O}}
\newcommand{\DiscreteMatching}{\eta(^{\bfx}\hat{p}^{[d]})}
\newcommand{\DiscreteTrj}{\eta(^{\bfx}p^{[d]})}
\newcommand{\DTWdist}[2]{||#1||_{DTW}^{#2}}

\stackMath
\newcommand\reallywidehat[1]{%
\savestack{\tmpbox}{\stretchto{%
  \scaleto{%
    \scalerel*[\widthof{\ensuremath{#1}}]{\kern-.6pt\bigwedge\kern-.6pt}%
    {\rule[-\textheight/2]{1ex}{\textheight}}
  }{\textheight}%
}{0.5ex}}%
\stackon[1pt]{#1}{\tmpbox}%
}
\newcommand{\defeq}{\vcentcolon=}

\newcommand{\objective}{baseline effectiveness~}

\theoremstyle{remark}
\newtheorem{remark}{Remark}

\begin{document}
\title{\LARGE \bf How to Evaluate Proving Grounds for Self-Driving? \\ A Quantitative Approach }

\author{Rui Chen,
        Mansur Arief,
        Weiyang Zhang,
        and~Ding Zhao
\thanks{This project
is funded in part by Carnegie Mellon University’s Mobility21 National University Transportation Center,
which is sponsored by the US Department of Transportation.}
\thanks{Rui Chen, Mansur Arief, Weiyang Zhang  and Ding Zhao are with the Department of Mechanical Engineering, Carnegie Mellon University, Pittsburgh, PA, 15232 USA e-mail: \{ruichen2, marief, weiyangz, dingzhao\}@andrew.cmu.edu.}

}


\maketitle
\IEEEpeerreviewmaketitle

\begin{abstract}
Proving ground has been a critical component in testing and validation for Connected and Automated Vehicles (CAV). Although quite a few world-class testing facilities have been under construction over the years, the evaluation of proving grounds themselves as testing approaches has rarely been studied. In this paper, we present the first attempt to systematically evaluate CAV proving grounds and contribute to a generative sample-based approach to assessing the representation of traffic scenarios in proving grounds. Leveraging typical use cases extracted from naturalistic driving events, we establish a strong link between proving ground testing results of CAVs and their anticipated public street performance. We present benchmark results of our approach on three world-class CAV testing facilities: Mcity, Almono (Uber ATG), and Kcity. We successfully show the overall evaluation of these proving grounds in terms of their capability to accommodate real-world traffic scenarios. We believe that when the effectiveness of a testing ground itself is validated, the testing results would grant more confidence for CAV public deployment.
\end{abstract}

\section{Introduction}
\IEEEPARstart{D}{evelopment} of self-driving technologies has drawn significant public attention with major concerns on safety issues. Before deploying connected and automated vehicles (CAVs), it is essential to test their performance both effectively and safely. Passchier \textit{et al.}~\cite{passchier_integral_2015} constructs a ``V-model" for the CAV development process. Szalay \textit{et al.}~\cite{szalay_structure_2016} visualized specifically the testing and validation process as a pyramid. Other testing methods have been proposed by intelligent vehicle researchers (Li~\textit{et al.}, \cite{li_intelligence_2016}; Koopman and Wagner, \cite{koopman_challenges_2016}; Zhao~\textit{et al.}, \cite{zhao_accelerated_2015}). Despite these extensive efforts on developing and testing CAVs, accidents still happen due to unverified faulty functionality and unexpected rare events. Such failure reveals both the imperfection of self-driving algorithms and the ineffectiveness of testing approaches. An effective testing process should be able to challenge CAVs with as many test cases from component level (e.g., sensor robustness) to scenario level (e.g., navigation through heavy traffic) as possible, and expose problems in advance~\cite{koopman_challenges_2016, amersbach_functional_2017}. More importantly, the testing process itself needs to be evaluated and verified in terms of testing capability so that one can anticipate CAV's driving performance from those testing results.

\begin{figure}[ht!]
    \centering
    \includegraphics[width=\columnwidth]{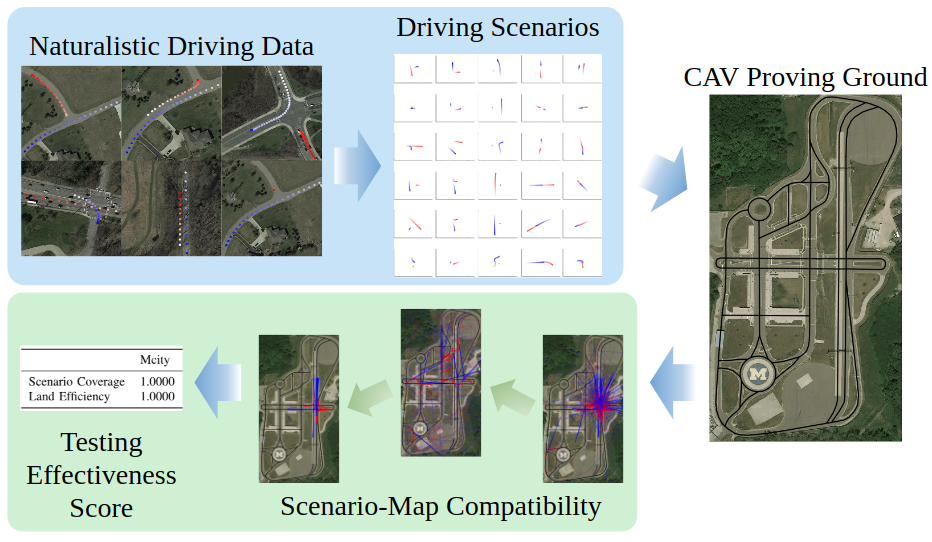}
     \caption{
     Driving scenarios are extracted from naturalistic driving data (shaded blue) and used for proving ground evaluation (shaded yellow).
    }
    \label{fig:highlevel}
\end{figure}

Regarding the CAV testing and validation pyramid \cite{szalay_structure_2016}, the two bottom levels, simulation and laboratory testing, are fully configurable and well-defined at the cost of testing fidelity and system integration. The top two levels, restricted public road and public road testing, serves as direct measure of CAV driving performance but are prone to the problem of insufficient encounter of rare and risky events. The middle level proving ground \cite{USDOT}, on the other hand, simulates real-world traffic environment in a reserved area with fully configurable physical assets and driving scenarios. Thus, proving grounds provide an integrated test method with high fidelity while having controllable testing risk. 

Traditionally, proving ground has been widely used to test performance of vehicles, for instance emission level, vehicle dynamics, advanced driver-assistance systems, design durability, etc. The long history of automobile industries and the testing requirement for an automobile product make conventional proving grounds widely available. Most of the conventional proving ground designs, however, are designed to certain functionalities \cite{van2016european, lee1997validation} and are not equipped with the capability of automotive to test various driving scenarios or behavioral competencies relevant for CAV safety evaluation \cite{ABCMcity}. Several CAV proving grounds \cite{mcity, castle, kcity, almono, trcsmart, harman} are built to incorporate these additional CAV testing capabilities. Since massive investment is required to build these testing facilities, an evaluation procedure will be of great benefit to systematically and quantitatively assess their effectiveness to validate the CAV functionalities. 

To date, we can not find work in the literature that assess the capability of proving grounds. Related work, for instance \cite{szalay2018development}, provides conceptual criteria to assess the effectiveness of a proving ground based on a focused group discussion involving academia, industries, and government officials. In the work, it is prescribed that a CAV proving ground should have several components, including urban tracks, complex city road elements, rural roads, highway sections, weather manipulations, signal-jamming areas, etc. Other work, for instance \cite{ABCMcity}, suggests the testing should be capable of recreating common-case scenarios and challenging or corner-case scenarios based on a certain scenario lists, which can be manually defined or extracted from large-scale driving data, to replicate various driving situations (e.g. weather, road conditions, etc) or behaviors of human drivers. A reasonable metric can therefore be derived by assessing how many scenarios a proving ground design can recreate and how fit the recreated scenarios are compared to their naturalistic occurrences in the real world. One way to achieve this is to use high-fidelity simulators, such as VSimRTI \cite{SCHUNEMANN20113189} and Nvidia drive constellation. An alternative approach directly relies on field study to gather human assessment of self-driving tests \cite{LIU2019354}. However, none of the existing work provides a provable quantitative measure for proving ground capabilities.


Notably, one challenge in quantitatively evaluating proving grounds is the difficulty in analytically representing traffic scenarios; they include emergent behaviors and patterns from the interactions between CAVs and the environments. Thus, in this work, we rely on a simplified scenario representation and derive a metric that evaluates the performance lower bound of proving grounds instead of their potential testing capabilities. We refer to this metric as \textit{baseline effectiveness} hereafter.

We evaluate the baseline effectiveness of a CAV proving ground by its capability of accommodating small use cases. In the literature, testing scenarios for autonomous driving are defined manually \cite{huang_task-specific_2014, nowakowski_development_2015,nhtsa_automated_2017}, based on heuristic rules (e.g. typical cases, corner cases, or dangerous cases) \cite{watzenig_automated_2017, li_intelligence_2016}, or derived from intelligence architecture perspective \cite{li_artificial_2018}. Due to the highly complicated and stochastic nature of real traffic, however, it is hard to empirically identify such a set that summarizes real-world traffic; driving scenarios could have uneven frequency of occurrence and various duration. An empirical set would be biased toward common scenarios while the rare or less intuitive ones should also be identified and considered just as essential. Therefore, it is necessary to extract a scenario set from naturalistic traffic data as our \textit{evaluation reference}. Each instance of the set should be a representation of one particular type of driving scenario with equally assigned significance. The set then serves as a summary of recorded traffic behaviors. 
In this paper, we focus on scenarios where vehicles interact in proximity and generate statistically inferential trajectories. Specifically, we compose the evaluation reference with \textit{traffic primitives} \cite{wang_extracting_2018, wang_clustering_2018} that are extracted from naturalistic driving data using sticky HDP-HMM methods as fundamental building blocks of multi-vehicle traffic scenarios. The main idea is that a set of discrete traffic primitives can be composed to form a large variety of driving behaviors.

Given an extracted traffic scenario set, we propose an evaluation algorithm based on particle filtering that assesses the target proving ground's road structure compatibility with the trajectories of the extracted traffic scenarios. Road structures is considered as the most common and essential attribute of both CAV proving ground and human driving scenarios, and thus used as the major representation in our evaluation process. One challenge to find the compatibility between vehicle trajectories and the test road is that we have no prior knowledge about the exact part of proving ground that should be used in such evaluation. In other words, we need to identify a road structure portion that best accommodates the traffic scenario first. The problem can then be equivalently formulated as a \textit{global localization problem} which positions a traffic scenario in the proving ground and maximizes its feasibility regarding the surrounding roads. To tackle this challenge, we propose a generative sample-based optimization method to simultaneously address the above two problems by finding a placement distribution of the traffic scenario. We develop a likelihood function based on Dynamic Time Warping \cite{salvador_toward_2007} (DTW) distance to assess the geometry similarity between target trajectories and proving ground road structure. Finally, we apply our approach on three state-of-the-art CAV proving grounds and assess their effectiveness in terms of two metrics: scenario coverage and land efficiency. See Fig. \ref{fig:highlevel} for an overview of our approach.

The rest of the paper is structured as follows. In Section \ref{sec:formulation}, we formally define our optimization problem, followed by an elaboration of scenario extraction and clustering method in Section \ref{sec:scenario_extraction} and a proposed solution in Section \ref{sec:WholeAlgDescription}. In Section \ref{sec:experiments}, we present evaluation results on selected CAV proving grounds. Section \ref{sec:conclusion} concludes this work with future directions.

\section{Problem formulation}\label{sec:formulation}

Notably, one major challenge in proving ground design is the immutability of constructed physical roads. Unlike other elements in CAV proving grounds, such as traffic facilities and dynamic road users, physical roads are rarely re-configurable once constructed. Thus, the testing road structure puts hard limitations on the possible test cases since real-world driving events implicitly include the specific surrounding road structure as an essential attribute. For instance, an high-speed turning can only happen at a long and curvy portion of a freeway. Consequently, the road map can be regarded as a \textit{proxy} for evaluating the effectiveness of proving grounds considering equal the potential of applying other elements to create high-fidelity traffic environments. We refer to testing traffic scenarios that are enabled by a CAV proving ground's road structure as the \textit{baseline effectiveness}. Due to the infeasiblity of complete testing on CAV systems \cite{koopman_challenges_2016} and evaluation directly based on the whole road map, it is natural for one to break enduring CAV tests into a variety of small and distinguished use cases, a.k.a typical \textit{scenarios} for autonomous driving \cite{wachenfeld_use_2016}, which will be explained next.

\subsection{Extraction of traffic scenario set}

We compose the reference for CAV proving ground evaluation by extracting traffic primitives from a set of $M$ naturalistic driving events $\bfY\defeq\{\rmY^{[m]}\}_{m=1}^M$ that involve $D$ vehicles. With $\bfs^{[m,d]}_t$ denoting state of the $d^{th}$ vehicle in event $m$ at frame $t$, a full data frame can be written as
\begin{equation}\label{def:dataframe}
    y^{[m]}_t\defeq<\bfs^{[m,d]}_t>_{d=1}^D,~\forall m\in[1,M]~\forall t\in[1,T_m]
\end{equation}

The driving event is then denoted as a time series
$$\rmY^{[m]}\defeq(y^{[m]}_t)_{t=1}^{T_m},~\forall m\in[1,M].$$
Alternatively we write a recorded event as the set of whole vehicle trajectories
$$\rmY^{[m]}\defeq\{y^{[m,d]}\}_{d=1}^{D},~\forall m\in[1,M].$$
where $y^{[m,d]}\defeq (\bfs^{[m,d]}_t)_{t=1}^{T_m}$ denotes the full trajectory of the $d^{th}$ vehicle in the $m^{th}$ event.

We would like to partition each event in $\bfY$ into individual driving scenarios. Specifically, we extract a scenario set $\bfZ\defeq\{{\rmZ}^{[q]}\}_{q=1}^Q~s.t.$
$$~\forall q\in[1,Q],~\exists m\in[1,M]\Rightarrow {\rmZ}^{[q]} \underset{w.r.t.~t}{\overset{\mathrm{sub-sequence}}{\prec}}\rmY^{[m]}.$$
Notably, elements in $\bfZ$ do not overlap with each other. Then, we form the evaluation reference for CAV proving ground evaluation by clustering $\bfZ$ into $K$ categories, each of which representing one particular type of real-world driving scenario. The evaluation reference is finally denoted as $\bfC\defeq\{\bfC_k\}_{k=1}^K$, where $\bfC_k\defeq\{{\rmZ}^{[q]}\in\bfZ~|~c(q)=k\}$. $c(\cdot)$ denotes the classification function. See Appendix~\ref{table:denotation} for a summary of notations to represent traffic scenarios.

\subsection{Testing capacity of CAV proving grounds}

As mentioned previously, we evaluate a CAV proving ground by estimating the compatibility between its road structure with the events in evaluation reference. We treat such assessment as a localization problem; we try to find the best placement for each reference event in the target proving ground where the portion of road structure in proximity best fits the vehicle trajectories in that event. Notably, when searching for an optimal placement pose, we treat the multi-vehicle trajectory $\rmZ^{[q]}\in\bfZ$ as a rigid body and preserve the relative orientation of whole trajectories $(z^{[q,1]},z^{[q,2]},\dots,z^{[q,D]})$ for all $D$ vehicles in $\rmZ^{[q]}$.

Additionally, we define the road structure of a CAV proving ground as $\bfR\defeq\{\rmR^{[i]}\}_{i=1}^{|\bfR|}$, where $\rmR^{[i]}$ refers to the $i^{th}$ road, characterized by a sequence of knots $r$:
$$\rmR^{[i]}\defeq(\bfr^{[i]}_j)_{j=1}^{|\rmR^{[i]}|},~\forall i\in[1,|\bfR|]$$
$\bfr^{[i]}_j\in\bbR^2$ denotes the coordinate of the $j^{th}$ knot in $i^{th}$ road.

Given the evaluation reference $\bfC$, we then define the \objective $\eff\defeq\{\efff{k}\}_{k=1}^K$ of a CAV proving ground with road structure $\bfR$ as the expected maximum placement probability within each scenario category:
\begin{maxi}|s|
{\bfx\in\bfG}{p(\bfx|\bfR, \rmZ)\Bigg],~\forall k\in[1,K]}
{}{\efff{k}=\bfE_{\rmZ\in\bfC_k}\Bigg[}\label{def:mapCapacity}
\end{maxi}
where $\bfx\defeq<t_x, t_y, \theta>$ denotes the 2D transformation pose of target scenario trajectory within boundary $\bfG$.


\section{Scenario extraction and clustering}\label{sec:scenario_extraction}
The construction of our evaluation reference leverages the work of Wang~\textit{et al.}~\cite{wang_understanding_2018} on extracting and clustering driving primitives from naturalistic driving data. The main idea is to treat human driving data as observations generated from a sequence of hidden \textit{patterns} which change on a larger time scale; human drivers make decisions less frequently than the actual trajectories are recorded. The human driving style in each of a short continuous period should remain stationary and thus leads to a specific traffic pattern. Traffic patterns are jointly determined by the traffic environment, vehicle state, and driving sub-goals, e.g., keeping in lane, making lane change, and doing a fast passing. In this paper, we formally define a scenario as a statistically inferable pattern consisting of continuous multi-vehicle data frames.

Scenario extraction by traffic patterns naturally serves as the fundamental building blocks of human driving since they represent typical driving behaviors if viewed separately, and, if viewed as a sequence, compose complete and versatile driving trajectories. Thus, we model human driving process as a Hidden Markov Model, or HMM, to capture the underlying dynamics of driving patterns. Specifically, we treat driving events as observations and traffic patterns as hidden states. To accommodate unknown but finite and discrete patterns, we apply a Hierarchical Dirichlet Process, or HDP, as the prior for pattern transitions. Additionally, a parameter $\kappa$ is introduced to control self-transition of hidden states. Hereby, we construct and solve a sticky HDP-HMM \cite{wang_extracting_2018, fox_sticky_2011, wang_driving_2018} to extract driving scenarios by collecting continuous data frames which share the same hidden state. Extracted driving scenarios are further clustered into $K$ predefined categories and form evaluation reference $\bfC$. We refer readers to \cite{wang_understanding_2018} for a complete description and implementation of the sticky HDP-HMM model. We will introduce the essential formulation of driving primitive extraction below. For more details on traffic primitives, please see \cite{wang_extracting_2018}.

\subsection{Human driving as HMM}
We denote the hidden states in HMM as $x_t\in\mathcal{X}$, which symbolically represent the traffic patterns appearing in driving events. Traffic pattern transitions are characterized by transition matrix $\pi$ where $\pi_{i,j}$ denotes the transition probability from state $i$ to $j$, i.e., $$x_t\sim\pi_{x_{t-1}},~\textrm{where}~\pi_{x_{t-1}}\defeq[\pi_{x_{t-1},1},~\pi_{x_{t-1},2},~\dots]$$
Observation $y_t$ is generated by the emission function $F$, i.e.,
$$y_t\sim F(\theta_{x_t})$$
where $\theta_x$ is the emission parameter associated with hidden state $x$, $\forall x\in\mathcal{X}$. Therefore, for a single driving scenario
all observations $z^{[q]}_t$ share the same hidden state, i.e., ${x}^{[q]}_{t_1} = {x}^{[q]}_{t_2} =\dots={x}^{[q]}_{t_{T_q}}$.


\subsection{HDP as prior for hidden state transition}
We apply Hierarchical Dirichlet Process to adaptively infer the number of hidden states, or the number of different traffic scenarios, in different driving events. Specifically, a discrete probability distribution $G_0$ is first drawn from a Dirichlet Process with parameter $\gamma$ and base measure $H$ as the common preference for all possible driving patterns

\begin{subequations}\label{eq:stickyHDPHMM}
\begin{equation}
G_0\sim DP(\gamma, H)
\tag{\ref{eq:stickyHDPHMM}}
\end{equation}
which is explicitly written with sampled patterns $\{\omega_i\}$ and stick-breaking weights $\{\beta_i\}$:
\begin{align}
    G_0 &= \sum_{i=1}^\infty\beta_i\delta_{\omega_i},~\omega_i\sim H\label{eq:commonPref}\\
    \beta_i &= v_i\prod_{\ell=1}^{i-1}(1-v_\ell),~v_i\sim Beta(1,\gamma)\label{eq:stickBreaking}
\end{align}
\end{subequations}
$\delta_\omega$ is a mass concentrated at $\omega$. Then, for each traffic pattern $i$, a transition distribution $\pi_i$ is sampled from a DP with $G_0$ as base measure and $\alpha$ as concentration parameter. Consequently, each transition distribution $\pi_i$ can be viewed as a variation of $G_0$. Finally, parameter $\kappa\in[0,1]$ is introduced to control the expected self-transition of traffic patterns, as shown in Eq.~\eqref{eq:transition}.
\begin{equation}\label{eq:transition}
    \pi_i\sim DP(\alpha+\kappa,\frac{\alpha\beta+\kappa\delta_i}{\alpha+\kappa}),~\forall i\in\mathcal{X}
\end{equation}

\section{CAV Baseline Effectiveness Measure}\label{sec:WholeAlgDescription}
In this section, we propose to use a particle filter-based algorithm to solve CAV proving ground effectiveness as defined in \eqref{def:mapCapacity}. Particle filtering \cite{gordon_novel_1993,kitagawa_monte_1996,doucet_sequential_1998} provides a computationally tractable solution for recursive Bayesian inference by maintaining a set of \textit{particles} that are drawn from the estimated distribution. The particles are recursively propagated according to a predefined motion model, weighted using a likelihood function, and resampled according to the assigned weights. It is particularly useful in dealing with nonlinear systems in that it does not impose any requirement on the form of motion and observation models. Such flexibility triggers a common usage of particle filters with adaption to specific situations. For instance in human motion tracking, McKenna~\textit{et al.}~\cite{mckenna_tracking_2007} applies particle filtering with iterated likelihood weighting on partial samples to recover from uncertain human motions and imperfect motion models. Notably, one key advantage of particle filtering in localization problems is its ability to represent multi-modal probability distributions \cite{dellaert_monte_1999}. Such capability prerequisites our approach in that there could be multiple parts of the CAV proving ground that accommodates a traffic scenario. Additionally, particle filtering allows nonlinear observation models and enables realistic compatibility measures for traffic scenario and road structures. First, we rewrite Eq.~\eqref{def:mapCapacity} as follows:
\begin{equation}\label{eq:ExpectedTauStar}
    \efff{k} = \bfE_{\rmZ\in\bfC_k}\Big[\scenecomp\Big],~\forall k\in[1,K]
\end{equation}
where
\begin{maxi}|s|
{\bfx\in\bfG}{\tgtdisNoIndex}
{}{\scenecomp=}\label{optim:placement}
\end{maxi}
We refer to $\scenecomp$ as the \textit{scenario compatibility} of $Z$ with $\bfR$ hereafter. We treat $\rmZ$ from each cluster as independent samples that jointly form the population of $\bfC_k$. Thus, we solve Eq.~\eqref{eq:ExpectedTauStar} by repeatedly solving the optimization problem defined by \eqref{optim:placement} with each sampled scenario $\rmZ$.
We proceed by viewing the problem as a \textit{global localization} of the scenario $\rmZ$ in proving ground $\bfG$ where the road structure best accommodates the trajectories. We assume that scenario $\rmZ$ is accommodated by a fixed part of the road structure $\bfR$ in area $\bfG$ if placed in a certain pose $\bfx^*$. Then, the maximization target $\tgtdisNoIndex$ in \eqref{optim:placement} is treated as the posterior distribution of placement pose $x$ of traffic scenario $\rmZ$ after observing the CAV proving ground $\bfR$. We then view the target placement as an unobserved static-state Markov process with the proving ground road structure as its observation. We assume that the state is subject to a dynamic process defined by a known stochastic motion model
\begin{equation}\label{def:motion_model}
\bfx_{t+1}=f_t(\bfx_t,w_t)    
\end{equation}
where $w_t$ denotes a noise with known statistics. We then define
\begin{equation}
 \PTransition\defeq p(\bfx_{t+1}|\bfx_t)=p(w_t)
\end{equation}
as the transition probability from $\bfx_t$ to $\bfx_{t+1}$, where $\bfx_{t+1}=f_t(\bfx_t,w_t)$. Additionally, we assume that the observed road structure $\bfR$ is generated from a stochastic observation model
\begin{equation}
    \bfR_t = \Obs_t(\bfx_t, \rmZ, v_t)
\end{equation}
where $v_t$ denotes a zero-noise with know statistics. We then define
\begin{equation}
    \LTwoIndex{t}{t}\defeq p(\bfR_{t}|\bfx_{t},\rmZ)=p(v_t)
\end{equation}
as the observation likelihood function.\\

\begin{remark}[Motion Model]
Generally, the motion model takes action $u_t$ as a parameter and is written as $f_t(\bfx_t, u_t, w_t)$. As such, the noise $w_t$ models imperfect motion sensor feedback, such as noisy encoder reading \cite{durrant-whyte_simultaneous_2006, cadena_past_2016}. Under the static state assumption in our case, the randomness of state transition solely comes from the fact that the true state is initially unknown. Noise $w_t$ drives the exploration in state space $\bfG$ for an approximation of the target posterior $\tgtdisNoIndex$.\\
\end{remark}

\begin{remark}[Observation Model]\label{remark:observation}
In the general form of HMM, the observation model captures the stochastic observation given a certain unobserved state; e.g., we normally treat sensors measurement as distributed in a certain range within the theoretical value. In our case, The observation model maps scenario $\rmZ$ with pose $\bfx_t$ to the required road structure $\bfR$. It is rare, however, to find a road structure in the proving ground that exactly match the scenario trajectories. Thus, we interpret noise $v_t$ as a distortion that leads to sub-optimal road structures; we allow road structure $\bfR$ to \textit{approximately} accommodate scenario $\rmZ$ with $v_t$ describing the discrepancy between them. The optimization objective is then embedded in the known statistics of $v_t$. In this paper, we assume that larger discrepancy is associated with a lower probability $p(v_t)$ and target at a solution of $\bfx$ that minimizes the discrepancy by pruning particles with low value of $p(v_t)=\LTwoIndex{t}{t}$. The probability of such discrepancy will be calculated by the likelihood function (see \ref{sec:likelihood}) which directly compares the corresponding road structure $\bfR_t$ and hypothesized placement pose $\bfx_t$, given the sampled scenario $\rmZ$.\\
\end{remark}

\begin{algorithm*}[!ht]
\caption{Modified Bayesian Bootstrap Filter for Maximizing Placement Probability of Single Scenario}
\label{alg:evalsingle}
\setstretch{1.1}
\begin{algorithmic}[1]
\Function{ComputeSingleScenario $\mathbf{e}_\bfR$}{$\rmZ$}
    \State Initialize $\{\bfx_0(i)\}_{i=1}^N$ by uniformly sampling from solution space $\bfG$
    \State $t\gets0$
    \Repeat
        \State Generate proposal samples $\{\bfx'_{t+1}(i)\}_{i=1}^N$ by $\bfx'_{t+1}(i)=f_t(\bfx_t(i),w_t(i))$, where $\{w_t(i)\}_{i=1}^N\overset{i.i.d}{\sim} p(w_t)$
        \State Assign weights $q'_{t+1}(i)\gets\mathcal{L}(\bfR|\bfx'_{t+1}(i), \rmZ)$
        \State Sort sample proposal $\{\bfx'_{t+1}(i)\}_{i=1}^N$ in ascending likelihood order such that $q'_{t+1}(i)\leq q'_{t+1}(j),~\forall i<j$
        \State Compute normalized weight $\widebar{q}'_{t+1}(i)\gets\left. q'_{t+1}(i)\middle/\sum_{i=1}^Nq'_{t+1}(i)\right.$
        \For{$i=1$ {\bf to} $\floor*{\rho_{r}\cdot N}$}
            \State Generate $\bfx_{t+1}(i)$ such that $P\big\{\bfx_{t+1}(i)=\bfx'_{t+1}(j)\big\}=\widebar{q}'_{t+1}(j),\forall j$\Comment{Resample low-likelihood samples}
        \EndFor
        \For{$i=\floor*{\rho_{r}\cdot N}+1$ {\bf to} $N$}
            \State $\bfx_{t+1}(i)=\bfx'_{t+1}(i)$\Comment{Preserve high-likelihood samples}
        \EndFor
        \State Calculate mean likelihood of the new samples $\widebar{q}_{t+1}=\frac{1}{N}\sum_{i=1}^N\mathcal{L}(\bfR|\bfx_{t+1}(i),\rmZ)$
        \State Store most likely sample $\{\bfx_{t+1}^*,q_{t+1}^*\}$
        \State $t\gets t+1$
    \Until{$t\geq T_0$ {\bf or} $q^*_t\geq \tilde{q}$ {\bf or} $\widebar{q}_t\geq \rho_c\cdot q_{t}^*$}\label{algl:converge}
    \State $T\gets t$
    \State\Return $\{\bfx_{T}^*,q_{T}^*\}$
\EndFunction
\end{algorithmic}
\end{algorithm*}

Applying Bayesian recursive estimation \cite{gordon_novel_1993}, we write the update rule for estimating $\tgtdisNoIndex$ as Eq.~\eqref{eq:bayesianRecursive}.
\begin{align}\label{eq:bayesianRecursive}
    &p(\bfx_{t+1}|\bfR_{1:t+1},\rmZ)\nonumber\\
    =~&\frac{\LTwoIndex{t+1}{t+1}p(\bfx_{t+1}|\bfR_{1:t},\rmZ)}{p(\bfR_{t+1}|\bfR_{1:t},\rmZ)}\nonumber\\
    \propto~&\LTwoIndex{t+1}{t+1}\int_{x_{t}}\PTransition\underbrace{p(\bfx_{t}|\bfR_{1:t},\rmZ)}_\textrm{Previous Estimation}d\bfx_t
\end{align}
Since we assume an HMM with static state and observation, $\bfR_{1:t}$ reduces to $\bfR$. Only the belief of $\bfx_t$ is changing with each recursive update. Thus, Eq.~\eqref{eq:bayesianRecursive} reduces to the actual update rule we use in our optimization approach as shown in Eq.~\eqref{eq:actualUpdateRule}.
\begin{subequations}\label{eq:actualUpdateRule}
\begin{equation}\tag{\ref{eq:actualUpdateRule}}
    p(\bfx_{t+1}|\bfR,\rmZ) \propto \LXIndex{t+1}\ProposalDist
\end{equation}
$\ProposalDist$ denotes the proposal distribution, or $prediction$ of $\bfx_{t+1}$ as shown in Eq.~\eqref{eq:proposal}.
\begin{equation}\label{eq:proposal}
    \ProposalDist=\int_{\bfx_{t}}\PTransition\tgtdis{t}d\bfx_t
\end{equation}
\end{subequations}

We then implement a modified Bayesian bootstrap filter \cite{gordon_novel_1993} to estimate the target distribution $\tgtdisNoIndex$. We maintain a finite discrete set $\{\bfx_t(i)\}_{i=1}^{N}$, or \textit{particles}, to approximate $\tgtdis{t}$ at iteration $t$. At iteration $t+1$, we pass each particle $\bfx_t(i)$ from iteration $t$ to the motion model $f_t$ and get the predicted state $\bfx'_{t+1}(i)=f_t(\bfx_t(i), w_t(i))$, where $w_t(i)$ is a sample drawn from a known statistics $p(w_t)$. Each predicted sample is then assigned a weight according to the likelihood function $q_t(i)=\mathcal{L}(\bfR|\bfx'_{t+1}(i),\rmZ)$. Finally, we generate the particles $\{\bfx_{t+1}(i)\}_{i=1}^N$ to approximate $\tgtdis{t+1}$ by sampling from predicted set $\{\bfx'_{t+1}(i)\}$ with probabilities proportional to $\{q_t(i)\}_{i=1}^N$, i.e., $\forall i,~P\{\bfx_{t+1}(i)=\bfx'_{t+1}(j)\}\propto q_t(j)$. Initial samples are drawn uniformly from solution space $\bfG$ for each attribute.

The algorithm recursively propagates particles across the solution space, updates them, and prunes erroneous ones with low-likelihood. The algorithm terminates when any of the following happens: 1) a maximum number of iterations $T_0$ is reached, 2) the highest sample likelihood reaches a threshold $\tilde{q}$, and 3) all samples converge to a local maximum. The last condition is indicated by a the mean sample likelihood approaching the highest sample likelihood by a ratio $\rho_c\in(0,1)$ (see Algorithm~\ref{alg:evalsingle} line \ref{algl:converge}). After convergence at iteration $T$, the optimal target $\scenecomp$ is approximated by the weight of the most likely particle $\bfx_T^*$, i.e.,
\begin{equation}
\begin{aligned}
    \widehat{\bfx}^*=\bfx_T^*&=\textrm{argmax}_{\bfx_T(i)}\mathcal{L}(\bfR|\bfx_T(i),\rmZ)\\
    \scenecomphat&=\mathcal{L}(\bfR|\widehat{\bfx}^*,\rmZ)=q^*_T
\end{aligned}
\end{equation}
See Algorithm~\ref{alg:evalsingle} for the complete process of computing the compatibility of a single traffic scenario $\rmZ$ with road structure $\bfR$. For each scenario cluster $\bfC_k$, we repeat the above procedure for every scenario $\rmZ$ in $\bfC_k$, and calculate \objective with respect to the $k^{th}$ scenario type as
\begin{equation}\label{def:objectivek}
    \efff{k}=\frac{1}{|\bfC_k|}\sum_{\rmZ\in\bfC_k}\scenecomphat
\end{equation}
See Algorithm.~\ref{alg:wholealg} for the workflow of evaluating CAV proving ground $\bfR$ with evaluation reference $\bfC$. The data preprocessing will be discussed in Section.~\ref{sec:data}.

\begin{algorithm}
\caption{CAV proving ground evaluation}
\label{alg:wholealg}
\setstretch{1.1}
\begin{algorithmic}[1]
\Function{BaselineEffectiveness}{$\bfC$,$\bfR$}
    \State \Call{Preprocessing}{$\bfC$,$\bfR$}\label{algl:prep}
    \For{$k=1$ {\bf to} $K$}
        \State $\widehat{\mathcal{E}}_k\gets 0$
        \ForAll{$\rmZ$ {\bf in} $\rmC_k$}
            \State $\widehat{\mathcal{E}}_k\gets\widehat{\mathcal{E}}_k+\scenecomp$ \Comment{see Algorithm~\ref{alg:evalsingle}}
        \EndFor
        \State $\widehat{\mathcal{E}}_k\gets\widehat{\mathcal{E}}_k/|\rmC_k|$
    \EndFor
    \State\Return $\{\efff{k}\}_{k=1}^K=\{\widehat{\mathcal{E}}_k\}_{k=1}^K$
\EndFunction
\end{algorithmic}
\end{algorithm}

\subsection{Modified Bayesian bootstrap filter}\label{sec:whole_algorithm}
One major change we made to the recursive update process is that only partial samples with the lowest likelihood are resampled in each iteration. The major motivation is that, in global localization problems, the initial iterations should focus on exploring solution space instead of pruning low-likelihood particles. Given no prior knowledge of the optimal scenario placement, an aggressive convergence could fall into local maximum easily. Partial resampling allows most samples to explore within medium-likelihood area for potential high-likelihood poses, while "restarting" the worst particles allows exploration on a larger scale. McKenna~\textit{et al.}~\cite{mckenna_tracking_2007} also split the samples into two halves; one half is kept updated for high-likelihood region exploration while the other half updates much less frequently to take advantage of accurate priors and motion model. Our approach and theirs both update partial samples more often then the other. One major different is that, our approach still diffuse the preserved particles since we do not have prior knowledge to rely on, while they skip full iterations for these samples.

\subsection{Scenario-Map Compatibility}\label{sec:likelihood}
In this section, we describe the likelihood function used in the recursive update \eqref{eq:actualUpdateRule}. Given road structure $\bfR$ of a CAV proving ground and scenario $\rmZ$, the likelihood function $\LNoIndex$ measures the quality of a 2D transformation pose $\bfx$. As discussed previously, we use vehicle trajectories, the most transferable representation of real-world traffic, to evaluate the compatibility between transformed traffic scenarios and CAV proving grounds. In a traffic scenario $\rmZ$ uniformly sampled from $\bfZ$, we denote the whole trajectory for the $d^{th}$ vehicle as $p^{[d]}\defeq\{p^{[d]}_t\}_{t=1}^{|\rmZ|}$, where $p^{[d]}_t\in\bbR^2$ denotes the GPS coordinate of that vehicle at frame $t$ in homogeneous coordinates. With hypothesis pose $\bfx\defeq<t_x, t_y, \theta>$, we first construct a 2D rigid body transformation matrix $\rmT_\bfx$
\begin{equation}
    \rmT_\bfx\defeq
    \begin{bmatrix}
        \cos(\theta) & -\sin (\theta) & t_x \\
        \sin (\theta) & \cos (\theta) & t_y \\
        0 & 0 & 1
    \end{bmatrix}
\end{equation}
Then, we generate the trajectory with hypothesis placement $^{\bfx}p^{[d]}$ by transforming the vehicle's whole trajectory according to $\bfx$
\begin{equation}
    ^{\bfx}p^{[d]}_t=\rmT_\bfx\cdot p^{[d]}_t,\forall t
\end{equation}

Now we would like to examine how road structure $\bfR$ accommodates $^{\bfx}p^{[d]}$. Since $\bfR$ and vehicle trajectory $p^{[d]}$ are collected from real-world data independently and represented using a sequence of 2D coordinates, diverse resolutions are expected. Different geometric distances between consecutive points will bias the compatibility measure. Thus, we first unify their resolution by constructing an binary occupancy grid $\mathcal{M}$ over the proving ground boundary $\bfG$ and register both $\bfR$ and $^{\bfx}p^{[d]}$ with grid $\mathcal{M}$. In specific, we define a data unification function $\eta$ that maps an arbitrary 2D point sequence $\{p_n\}_{n=1}^N$ to a sequence of positions in grid $\mathcal{M}$ with the same length as shown in equations \eqref{def:eta}.
\begin{subequations}\label{def:eta}
\begin{equation}\tag{\ref{def:eta}}
    \eta:\{\bbR^2\}_N\rightarrow\{\bbR^2\}_N
\end{equation}
where
\begin{equation}
    \eta(p) = \left(\floor*{\left. p_n\middle/
    \begin{bmatrix}
        \lambda_x \\
        \lambda_y
    \end{bmatrix}\right.}\right)_{n=1}^N 
\end{equation}
$\lambda_x$ and $\lambda_y$ are grid sizes along $x$-aixs and $y$-axis. $\floor{\cdot}$ refers to rounding down.
\end{subequations}
Applying \eqref{def:eta} to both road structure $\bfR$ and transformed vehicle trajectory $^{\bfx}p^{[d]}$, we let $\eta(\bfR)$ and $\DiscreteTrj$ denote their unified representations with same resolution. Notably, $\eta(\bfR)$ is generated by applying $\eta$ to each road $\rmR^{[i]}$ in $\bfR$. 

\begin{figure}[]
    \centering
    \includegraphics[width=\columnwidth]{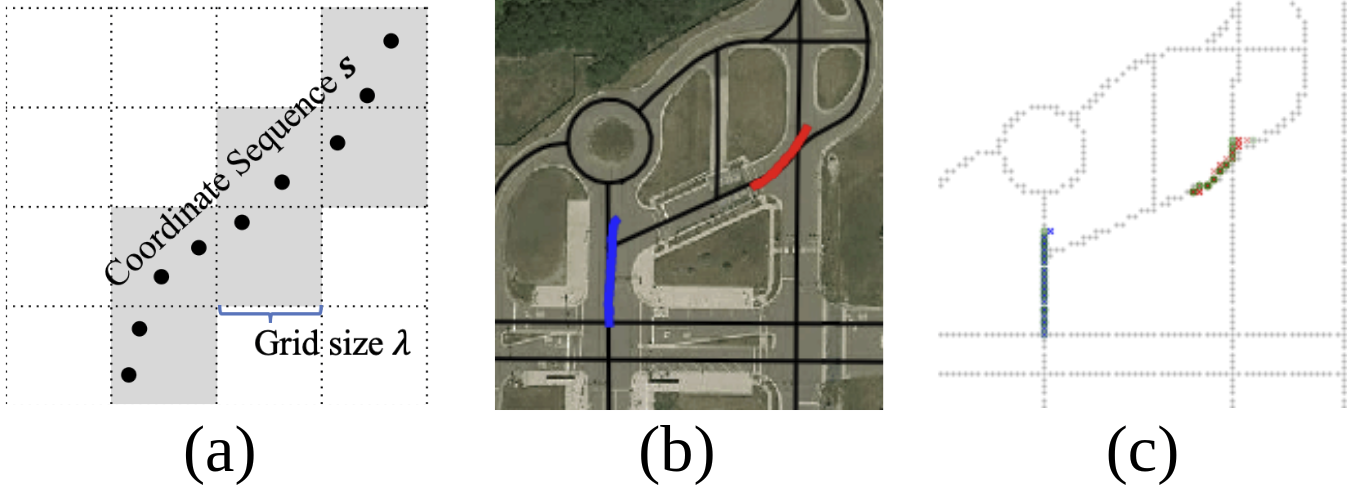}
     \caption{
     (a) We register each coordinate sequence $s$ with occupancy grid $\mathcal{M}$. (b) shows a sample traffic scenario (blue and red) and surrounding road structure (black). (c) For each occupied cell by vehicle trajectories (crosses in corresponding color), we find a nearest neighbor (green circle) among road cells (grey). We then compute the similarity between vehicle trajectory (crosses) and matching road segments (circles) as the scenario-map compatibility.
    }
    \label{fig:matching}
\end{figure}

Now, we evaluate the compatibility of trajectory $\DiscreteTrj$ and proving ground road structure $\eta(\bfR)$. Since we are searching for a placement of the trajectory $p^{[d]}$ where it is best accommodated, we can intuitively assess the compatibility between $\DiscreteTrj$ and $\eta(\bfR)$ by the proportion of $\DiscreteTrj$ that exactly overlaps $\eta(\bfR)$. However, we design a more flexible metric to allow \textit{approximate accommodation} with some distortion added to road structure $\bfR$ as discussed in Remark~\ref{remark:observation}. Notably, the scale of an extracted traffic scenario is often smaller than that of a CAV proving ground. Thus, we proceed by first extracting a segment of $\bfR$, or \textit{matching road segments}, that will most likely accommodate $\DiscreteTrj$; an intuitive choice is the collection of pixel-wise nearest neighbors of transformed trajectory $\DiscreteTrj$ in road structure $\eta(\bfR)$. Let $\DiscreteMatching$ denote the matching road segment of $^{\bfx}p^{[d]}$ in $\bfR$ registered with grid $\mathcal{M}$, we have
\begin{argmini}|s|
{r'\in\eta(\bfR)}{||r'-\eta_t(^{\bfx}p^{[d]})||_2,~\forall t\in[1,|\rmZ|]}
{}{\eta_t(^{\bfx}\hat{p}^{[d]})\defeq}\label{def:matching}
\end{argmini}
See Fig.~\ref{fig:matching} for an illustration of finding matching road segments.

\begin{figure*}[htbp!]
    \centering
    \includegraphics[width=\textwidth]{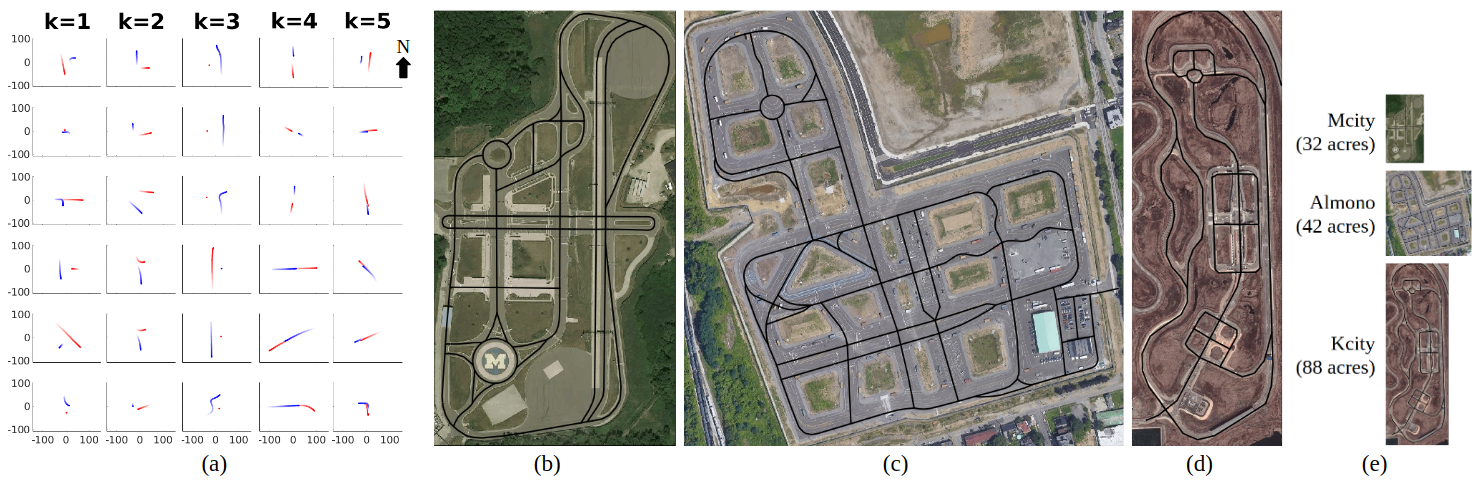}
     \caption{
     (a) shows examples of five traffic scenario categories we define (from left): $(k=1)$ at intersection with strong interaction, $(k=2)$ at intersection with light interaction, $(k=3)$ one vehicle static, $(k=4)$ on same lane with strong interaction, and $(k=5)$ on different lanes with light interaction. Trajectories of two vehicles are distinguished by red and blue with faded parts indicating earlier vehicle positions. All axis values are in meters. We collect GPS coordinates of road structures (marked in black) via openstreetmap and satellite image via Google Maps Static API for (b) Mcity, (c) Almono, and (d) Kcity. The satellite images are for visualization only and are not used in computation. See (e) for a size comparison.
    }
    \label{fig:database}
\end{figure*}

\begin{figure}[htbp!]
    \centering
    \includegraphics[width=\columnwidth]{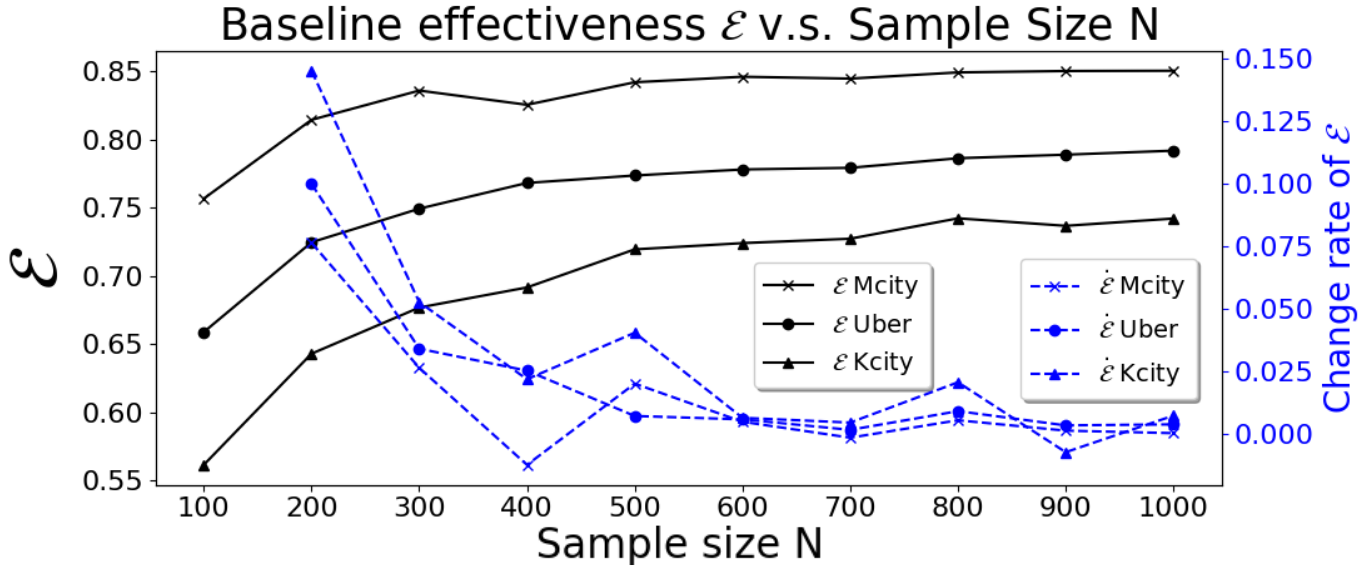}
     \caption{
     Baseline effectiveness of selected CAV proving grounds with random scenario set. The change rate with respect to sample size is also plotted, e.g., the effectiveness changes by $\sim 2\%$ if we increase the sample size from $700$ to $800$ for Kcity evaluation.
    }
    \label{fig:sample_size}
\end{figure}

\begin{figure*}[htp]
    \centering
        \includegraphics[width=\textwidth]{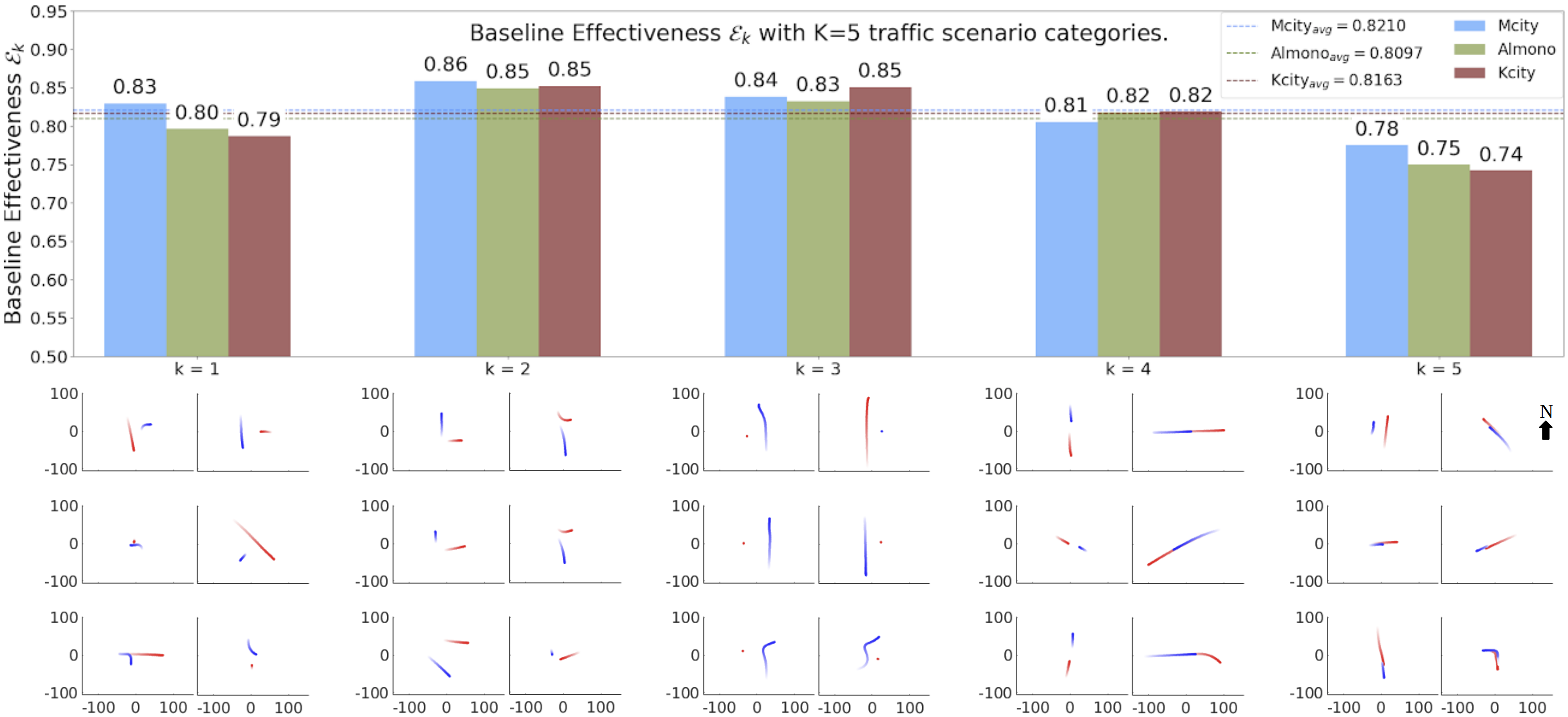}\label{fig:full_a}
        \vspace{0.1in}\footnotesize (a) Baseline effectiveness $\efff{k}$ of each CAV proving ground and example scenarios for each category.
        \includegraphics[width=\textwidth]{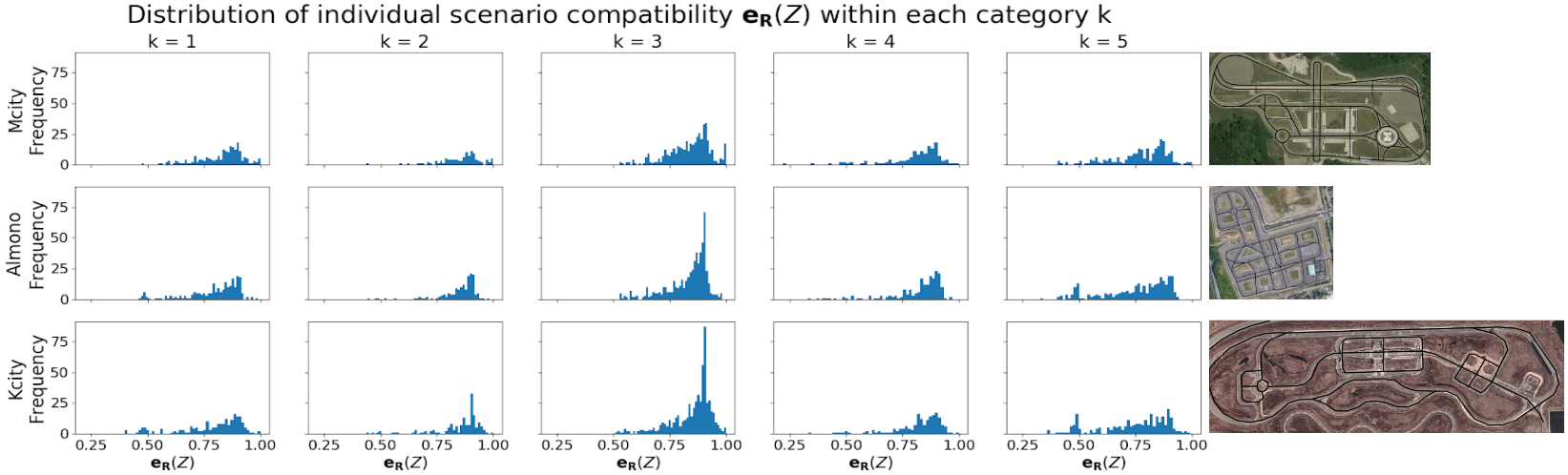}\label{fig:full_b}
        \vspace{0.1in}\footnotesize (b) Histogram of individual scenario compatibility.
    \caption{(a) shows the benchmark result of applying Algorithm~\ref{alg:wholealg} to Mcity, Almono, and Kcity. Scenario examples from each category is shown below the corresponding results. (b) shows the distribution of individual scenario compatibilities with each category evaluated for each CAV proving ground.}
    \label{fig:full}
\end{figure*}

\begin{figure*}[htp]
    \centering
        \includegraphics[width=\textwidth]{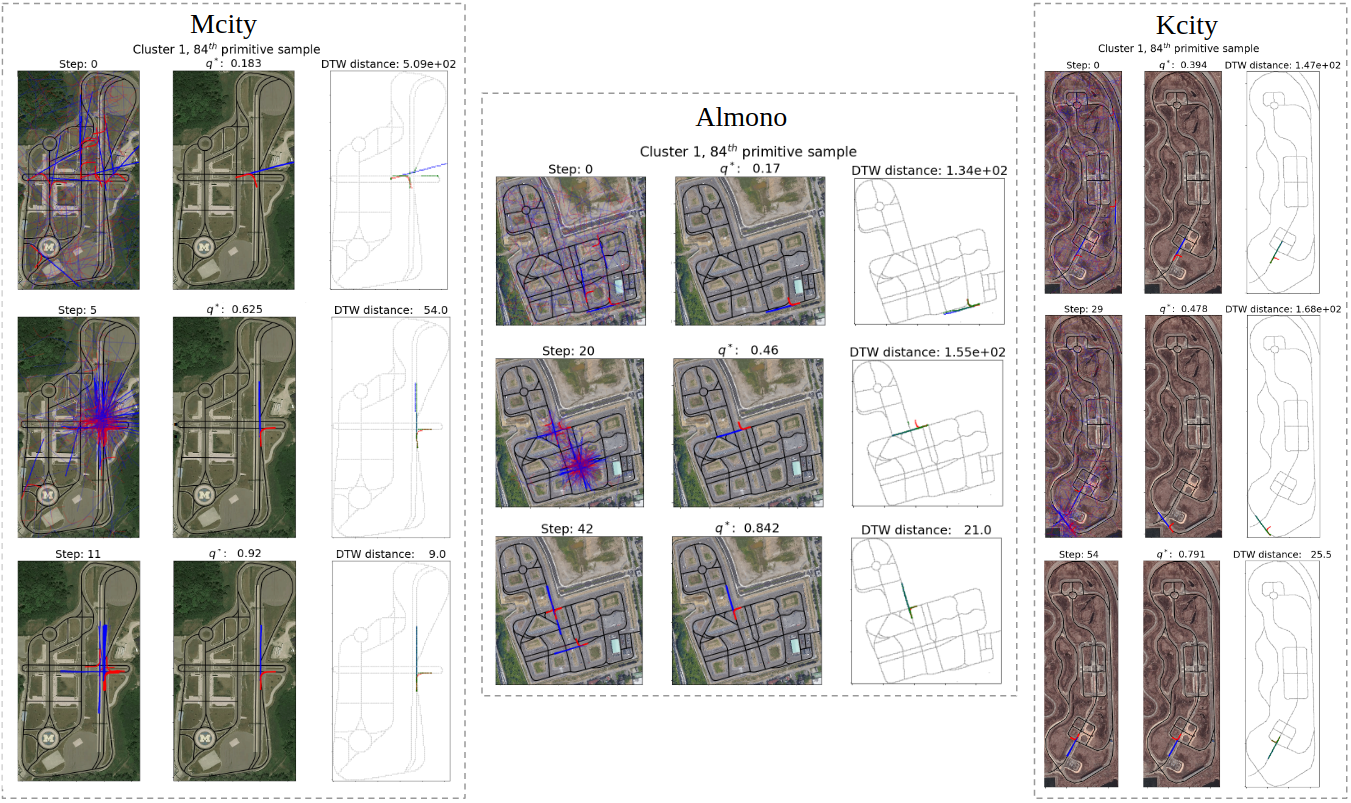}
        \vspace{0.1in}\footnotesize (a)
        \includegraphics[width=\textwidth]{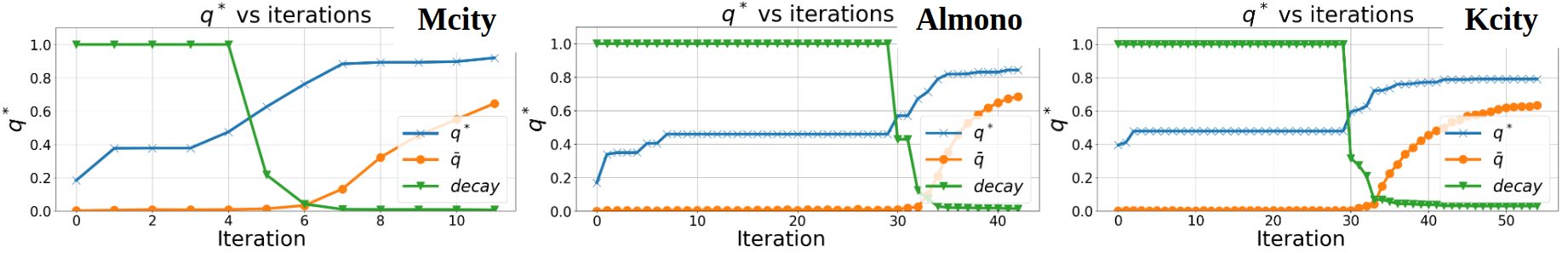}
        \vspace{0.1in}\footnotesize (b)
    \caption{(a) shows the computation of scenario compatibility of a sample scenario from category $1$ (at intersection with strong interaction) with Mcity, Almono, and Kcity. In each row of plot, all placement hypothesis (left), the most likely placement with weight in title (middle), and matching road segments for the most likely placement (right) are shown. (b) plots the weight of most likely placement (blue) during the recursive computation of scenario compatibility. The mean weight of all samples (orange) as well as the diffusion decay factor (green) are also plotted.}
    \label{fig:case1}
\end{figure*}

\begin{figure*}[htp]
    \centering
        \includegraphics[width=\textwidth]{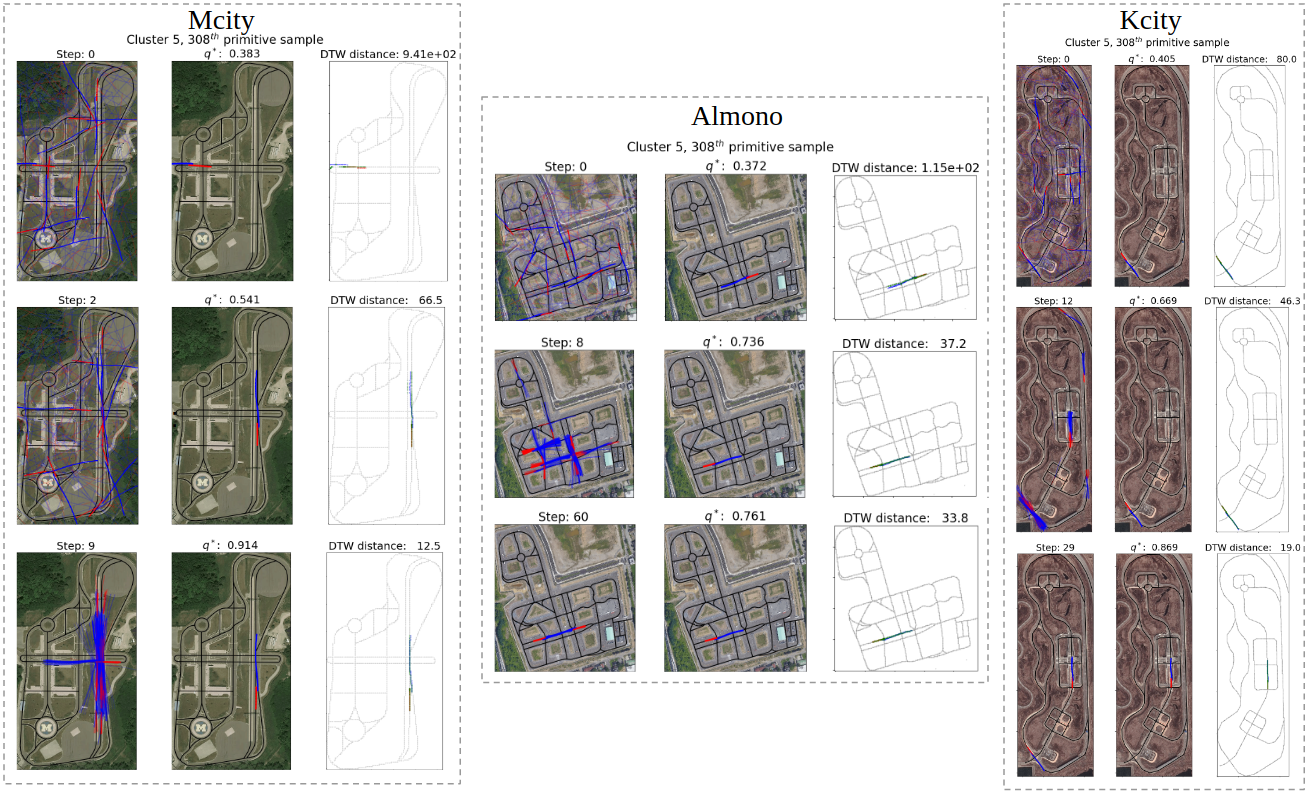}
        \vspace{0.1in}\footnotesize (a)
        \includegraphics[width=\textwidth]{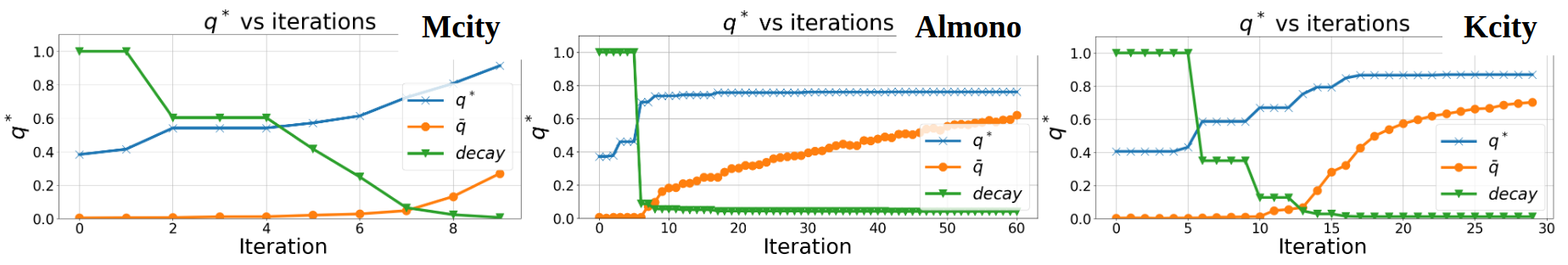}
        \vspace{0.1in}\footnotesize (b)
    \caption{The computation process of scenario compatibility of a sample scenario from category $5$ (on different lanes with light interaction) is shown in similar style to Fig.~\ref{fig:case1}.}
    \label{fig:case2}
\end{figure*}

Dynamic Time Warping \cite{salvador_toward_2007, ding_querying_2008, rakthanmanon_searching_2012} distance, or DTW distance, is used as the measure for sequence dissimilarity between the discrete transformed trajectory and its matching road segment. Given two sequences $X$ and $Y$ of length $|X|$ and $|Y|$, the DTW problem aims to find a warp path ${W=w_1,w_2,\dots,w_K}$, where $w_k=(i,j)$ denotes a path from the $i^{th}$ element in $X$ to the $j^{th}$ element in $Y$, such that the DTW distance $\DTWdist{X,Y}{W}$ \eqref{def:dtwdist} is minimized.
\begin{equation}\label{def:dtwdist}
        \DTWdist{X,Y}{W}=\sum_{k=1}^Kd(X(w_{ki}),Y(w_{kj}))
\end{equation}
We use Euclidean distance for $d(\cdot)$ in this paper since we are evaluating two sequences representing physical locations. Denoting the optimal warping path for minimizing DTW distance as $W^*$, we define \textit{DTW feasibility} $\xi\in[0,1]$ of the $d^{th}$ vehicle trajectory under placement pose $\bfx$ as
\begin{equation}\label{def:dtw_feasibility}
    \xi(p^{[d]}, \bfx, \bfR)=\left(1-\frac{\mid\mid\DiscreteTrj,~\DiscreteMatching\mid\mid_{DTW}^{W^*}}{|\DiscreteTrj|}\right)_{+}
\end{equation}
The measure defined above first finds the DTW dissimilarity between a transformed trajectory $p^{[d]}$ and its matching segment in road structure $\bfR$ registered with occupancy grid $\mathcal{M}$. The outcome is normalized by trajectory length to represent the proportion that fails to match. We then define DTW feasibility as the non-negative matching ratio. Finally, we define likelihood function as the mean DTW feasibility of all single trajectories in scenario $\rmZ$ under the same placement pose $\bfx$
\begin{equation}\label{def:likelihood}
    \LNoIndex\defeq\frac{1}{D}\sum_{d=1}^D\xi(p^{[d]},\bfx,\bfR)
\end{equation}

\section{Experiments and Discussion}\label{sec:experiments}

In this section, we describe the naturalistic driving data and the traffic scenarios extracted from them. Then, we perform experiments using models of CAV proving grounds to investigate the effectiveness of our approach and finally present evaluation results on real CAV testing facilities.

\subsection{Scenario Preparation and Proving Ground Data Collection}\label{sec:data}
We adopt the naturalistic driving data used by Wang \textit{et al.}~\cite{wang_understanding_2018}. The driving events are collected by the University of Michigan Safety Pilot Model Development (SPMD) program by University of Michigan Transportation Research Institute (UMTRI) \cite{bezzina_safety_2014}. The SPMD program deployed approximately $3,500$ vehicles equipped with Dedicated Short Range Communication (DSRC) in Ann Arbor, MI for more than 3 years and record vehicle trajectories when the distance between two vehicles are within $100$ meters. Totally $|\bfY|=976$ dual-vehicle driving events which last for more than $10$ seconds are extracted from SPMD database. Applying the sticky HDP-HMM as summarized in Section~\ref{sec:scenario_extraction}, we ended up with $4,126$ extracted traffic scenarios. We further select $|\bfZ|=2,476$ scenarios where at least one vehicle trajectory is longer than $5$ meters and cluster them into $K=5$ predefined categories. See Fig.~\ref{fig:database} for category definition and sample scenarios.

When choosing existing CAV proving grounds for demonstrating our approach, we select those that 1) are dedicated for testing self-driving technologies, 2) have road map logged on Open Street Map\footnote{Map data $\textcopyright$ OpenStreetMap contributors. Link to copyright page: https://www.openstreetmap.org/copyright} (OSM) and satellite image available via Google Maps Static API\footnote{https://developers.google.com/maps/documentation/maps-static/intro}, and 3) are not military facilities. Based on these criteria, we choose Mcity, Almono, and Kcity as our target CAV proving grounds. We export their road map $\bfR$ from OSM in GPS coordinates. In data preprocessing (see Algorithm~\ref{alg:wholealg} line~\ref{algl:prep}), for any geometry data $s\in\{\bfR,\rmZ^{[1]},\rmZ^{[2]},\dots,\rmZ^{[Q]}\}$ in GPS coordinates, we apply sinusoidal projection with mean longitude of $s$ as central meridian \cite{snyder_map_1987} to reduce distortion. 
See Fig.~\ref{fig:database} for example scenarios and projected proving ground road structures. All roads and trajectories are linearly interpolated to have a resolution of one meter.

\subsection{Implementation of Bayesian Bootstrap Filter}

As discussed in Section~\ref{sec:WholeAlgDescription}, the stochastic motion model \eqref{def:motion_model} drives the exploration of samples in the solution space to search for optimal placement. Due to static state assumption, the motion model becomes essentially a diffusion model. In this paper, we define the diffusion process as Gaussian diffusion
\begin{equation}
    \bfx'_{t+1}=f_t(\bfx_t,w_t) = \bfx_t+<w_{t,x}, w_{t,y}, w_{t,\theta}>
\end{equation}
where $w_{t,x}, w_{t,y}, w_{t,\theta}$ are sampled from zero-mean normal distributions with $\sigma_x,\sigma_y,\sigma_\theta$ as standard deviation, respectively. We take $\sigma_x=\sigma_y=8~m,~\sigma_\theta=\pi/2~rad$ in the experiments. Additionally, we decay the diffusion variances exponentially when the best sample weight exceeds $q_d=0.5$ to achieve a final convergence. Specifically, let $t$ denote the iteration index, we write the decay factor $\gamma_t$ at iteration $t$ as
\begin{equation}\label{def:vardecay}
    \gamma_t = \frac{1}{{1+\lambda_0t^2}}\left(\alpha^{q*_t-q_d}\right)^{\mathbb{I}[q^*_t\geq q_d]}
\end{equation}
where $\lambda_0=1e-3$ is the natural decay factor and $\alpha=5e-6$ is the exponential base. $\mathbb{I}[\cdot]$ refers to binary indicator function. Solution space $\bfG$ is defined as the smallest bounding box that closes all proving ground roads.

In resampling step, we take $\rho_r=0.6$ to preserve approximately the best $40\%$ particles in each update iteration. For algorithm termination, we take $\rho_c=0.8$ to indicate a convergence to local maximum when the mean sample likelihood reaches $80\%$ of the best sample likelihood; i.e., the sample population is so concentrated that improvement by exploration is no longer expected. The algorithm also terminates when the weight $q^*_t$ of most likelihood sample $\bfx_t^*$ reaches $\Tilde{q}=0.9$. The maximum iteration number is $T_0=300$. 

We take $\lambda_{x,y}=2m$ as the grid size for occupancy grid $\mathcal{M}$ along both $x$- and $y$-axis. When calculating DTW feasibility $\xi$ (see Eq.~\eqref{def:dtw_feasibility}), we use the python implementation of FastDTW \cite{salvador_toward_2007} to calculate the DTW distance $||\cdot,\cdot||_{DTW}^{W^*}$. The window parameter is not constrained.

\subsection{Benchmark Results on Existing CAV Proving Grounds}

In this section, we present the benchmark result of our approach on three of the world-class testing facilities dedicated for CAV: Mcity, Almono, and Kcity. Since we are using particle-based approximation for the placement distribution of traffic scenarios, the accuracy of such approximation is affected by the quantity of particles $N$
. As we increase the sample size $N$ to infinity, the estimated placement distribution should converge to the true value. Thus, we first run our evaluation approach with a random subset of traffic scenarios and keep increasing the sample size until 
the compatibility metric only changes by less than $1\%$. The final sample size we pick are 500, 500, and 800 for Mcity, Almono, and Kcity respectively. This choice is consistent with the fact that Kcity (88 acres) is approximately double the size of the other two facilities (32, 42 acres for Mcity and Almono respectively), leading to a significantly larger solution space. Then we perform the full baseline effectiveness benchmark on all selected proving grounds with all extracted traffic scenarios as shown in Fig.~\ref{fig:full}(a). The average compatibility tested on all scenarios are indicated by horizontal dashed lines. The bar plots show baseline effectiveness on each of the five scenario categories. We observe that all three proving grounds demonstrated similar overall testing capabilities with various specialties, which will be discussed in case studies. Fig.~\ref{fig:full}(b) shows a histogram of individual scenario compatibility $\scenecomp$ in each scenario category $\bfC_k$ with all three proving grounds. In all scenario category-map combinations, we observe normally distributed compatibility scores, whose arithmetic mean yields a feasible approximation $\reallywidehat{\mathcal{E}}_k$ to baseline effectiveness $\efff{k}$ as defined in Eq.~\eqref{def:objectivek}.

We also propose two metrics based on the previous benchmark results. First, we calculate the \textit{scenario coverage} score by computing compatibility score $\scenecomp$ among all driving scenarios. Scenario coverage evaluates the overall testing capability of a proving ground. Second, we calculate the \textit{land efficiency} by normalizing scenario coverage score by CAV proving ground size. The results are normalized with respect to Mcity scores and summarized in Table~\ref{table:eval}. We observe that despite occupying the least area, Mcity achieves the highest scenario coverage and yields the highest land efficiency.
\begin{table}[htbp]
  \centering
  \caption{Evaluation of CAV proving grounds}
    \begin{tabular}{llll}
    \toprule
     & Mcity & Almono & Kcity \\
    \midrule
    Scenario Coverage & 1.0000 & 0.9862 & 0.9943 \\
    Land Efficiency & 1.0000 & 0.7514 & 0.3616 \\
    \bottomrule
    \end{tabular}%
  \label{table:eval}%
\end{table}%

\subsubsection{Case study \#1}
\textit{Interaction with crossing trajectories.} As shown in Fig.~\ref{fig:full}, scenarios where the two vehicle trajectories are likely to intersect ($k=1,2$) are well-supported by Mcity, Almono, and Kcity. All three proving grounds achieves baseline effectiveness of $\sim 0.85$ when the interaction is light, since such scenario can be properly accommodated by a large variety of road structures as long as two roads intersect each other. Besides, all three proving grounds has less support for scenarios with strong vehicle interaction, since certain types of intersections are required. With a higher demand of vehicle interaction, Mcity has the best support because of its versatile road types (highway with ramps, intersections with various angles, roundabouts, etc.) despite that it has the smallest area. Other two proving grounds, on the other hand, provide less support since their road types are limited. In Almono, most blocks are square-shaped and and have approximately the same size. In Kcity, the versatility of intersections is comparable to those of Mcity. See Fig.~\ref{fig:case1} for a qualitative example. Notably, at iteration $5$, a satisfactory placement for the testing scenario is found in Mcity, yielding a sharp decay in diffusion variance, shortly before a placement with saturated weight ($0.92$) is found at iteration $11$. Finally, Mcity provides the best accommodation with a intersection formed near a freeway.
\subsubsection{Case study \#2}
\textit{Interaction with parallel trajectories.} Another important class of driving scenarios happens on the same road or parallel roads ($k=4,5$). Examples include highway platooning, merging, following, and driving in opposite directions. As shown in Fig.~\ref{fig:full}, all three proving grounds provides satisfactory support for single-lane interactions with at least $\sim 0.81$ baseline effectiveness score. Almono and Kcity performs slightly better than Mcity since they occupy more space and are more likely to accommodate long or high-speed scenarios. For scenarios where two vehicles driving in different lanes or roads, more flexible combinations of road structures are necessary. As such, all proving grounds provides less support while Mcity still performs the best. See Fig.~\ref{fig:case2} for qualitative examples. Mcity contains a ramp with very similar shape with the trajectories, and thus best accommodates the testing scenario. The lack of such road type in Almono and yields to a low score. In Kcity, although several highway merges and ramps exist, none of them creates the specific merging or splitting angle as required by testing scenario. However, Kcity still approximately support the scenario with a road segment with similar curves with desired trajectories.
\subsubsection{Possible improvements \#3}
We also identify some possible future work of our approach. First, the construction of evaluation reference and proving ground description with high-dimensional driving data is desired but remains a challenge. In this paper, we evaluate the effectiveness of CAV proving grounds based on static geometric information, a necessary yet general representation for driving scenarios. Under such setting, we apply a sticky HDP-HMM model to segment driving events and used predefined clustering metrics to construct the evaluation reference. Such method can be computationally intractable for high-dimensional driving data. Ideally, a traffic scenario should have a comprehensive list of attributes (road elevation, traffic facilities, traffic rules, etc.), more vehicles, dynamic information (e.g., trajectory feasibility considering real vehicle dynamics), and so on. However, it remains a challenge to model, segment, and cluster high-dimensional time sequences. Advances in these areas prerequisite the usage of a evaluation reference with higher fidelity, which will influence the evaluation results on selected testing facilities.

Second, a better road structure representation is desired. In this paper, we follow the road descriptor by OSM, where the physical information is simplified to nodes and links; i.e., lane number, road width, and elevations are neglected. With more descriptive representations (e.g., Lanelets, \cite{bender_lanelets:_2014}; Lanelet2, \cite{poggenhans_lanelet2:_2018}), the evaluation reference will capture real-world driving behaviors more realistically and lead to assessments with higher fidelity. In such case, the likelihood model should be adapted accordingly.

\section{Conclusions and Future work}\label{sec:conclusion}
In this paper, we have proposed an evaluation algorithm and metric for CAV proving grounds using a generative sample-based optimization approach. This paper presents the first attempt to systematically evaluate CAV proving grounds with respect to naturalistic logged driving data. Our approach directly utilizes the expected uses cases as the evaluation reference and hence provides solid connection between proving ground performance of CAVs and their expected public street performance. We present our evaluation approach on three world-class CAV proving grounds and evaluate their capability to accommodate real-world driving scenarios. Based on the evaluation results, we have quantitatively shown the overall testing capability of the three proving grounds as well as their land efficiency. CAV proving ground remains an essential, highly valuable, yet costly component of the validation of self-driving technologies. We believe that when the effectiveness of testing approach itself is verified, the corresponding testing performance brings a higher level of confidence for public street deployment of CAVs.


\ifCLASSOPTIONcaptionsoff
  \newpage
\fi

\begin{appendices}

\section{Notations for representing traffic scenarios}
\label{table:denotation}%
\begin{table}[htbp]
  \centering
    \begin{tabular}{ll}
    \toprule
    \multicolumn{1}{l}{Notation} & Meaning \\
    \midrule
    $\mathbf{Y}=\{\mathrm{Y}^{[m]}\}_{m=1}^M$ & Recorded driving scenario set.\\
    $m$ & Index of driving scenario.\\
    $d$ & Index of vehicle in driving scenarios.\\
    $y^{[m]}_t$ & Data frame of $\mathrm{Y}^{[m]}$ at time $t$.\\
    $y^{[m,d]}$ & Trajectory of vehicle $d$ in $\mathrm{Y}^{[m]}$.\\
    $s^{[m, d]}_t$ & State of vehicle $d$ in $\mathrm{Y}^{[m]}$ at time $t$.\\
    $\mathbf{Z}=\{\mathrm{Z}^{[q]}\}_{q=1}^Q$ & Extracted driving scenario set.\\
    \bottomrule
    \end{tabular}%
\end{table}%
\end{appendices}

\bibliography{references,ref}
\bibliographystyle{unsrt}




\begin{IEEEbiography}[{\includegraphics[width=1in,height=1.25in,clip,keepaspectratio]{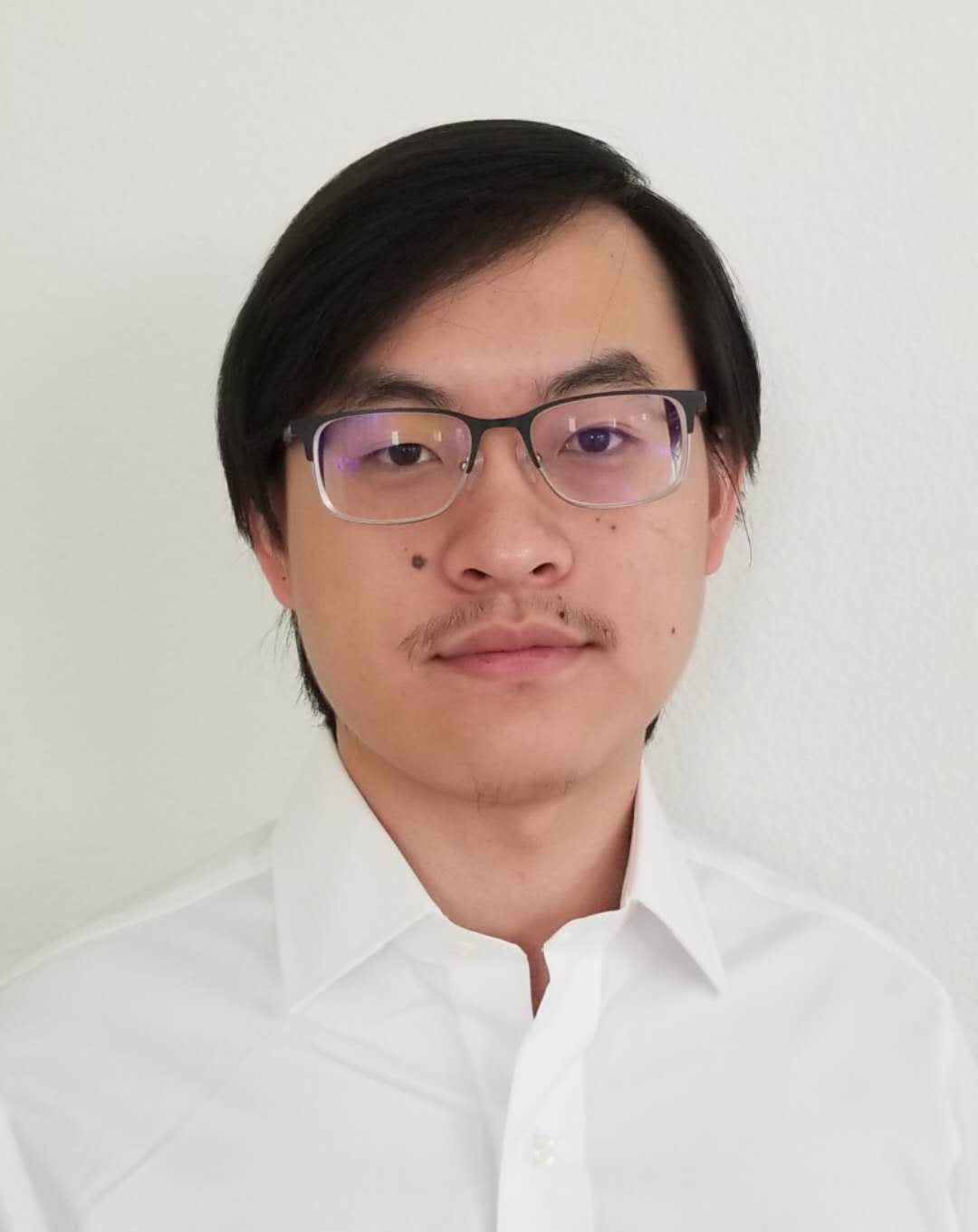}}]{Rui Chen} received his B.S. degree in 2017 and M.S. degree in 2018 from the University of Michigan, Ann Arbor with the department of Electrical Engineering and Computer Science. His research focuses on reinforcement learning, probabilistic robotics, and machine learning with applications in human-robot interaction, manufacturing, and autonomous driving.
\end{IEEEbiography}

\begin{IEEEbiography}[{\includegraphics[width=1in,height=1.25in,clip,keepaspectratio]{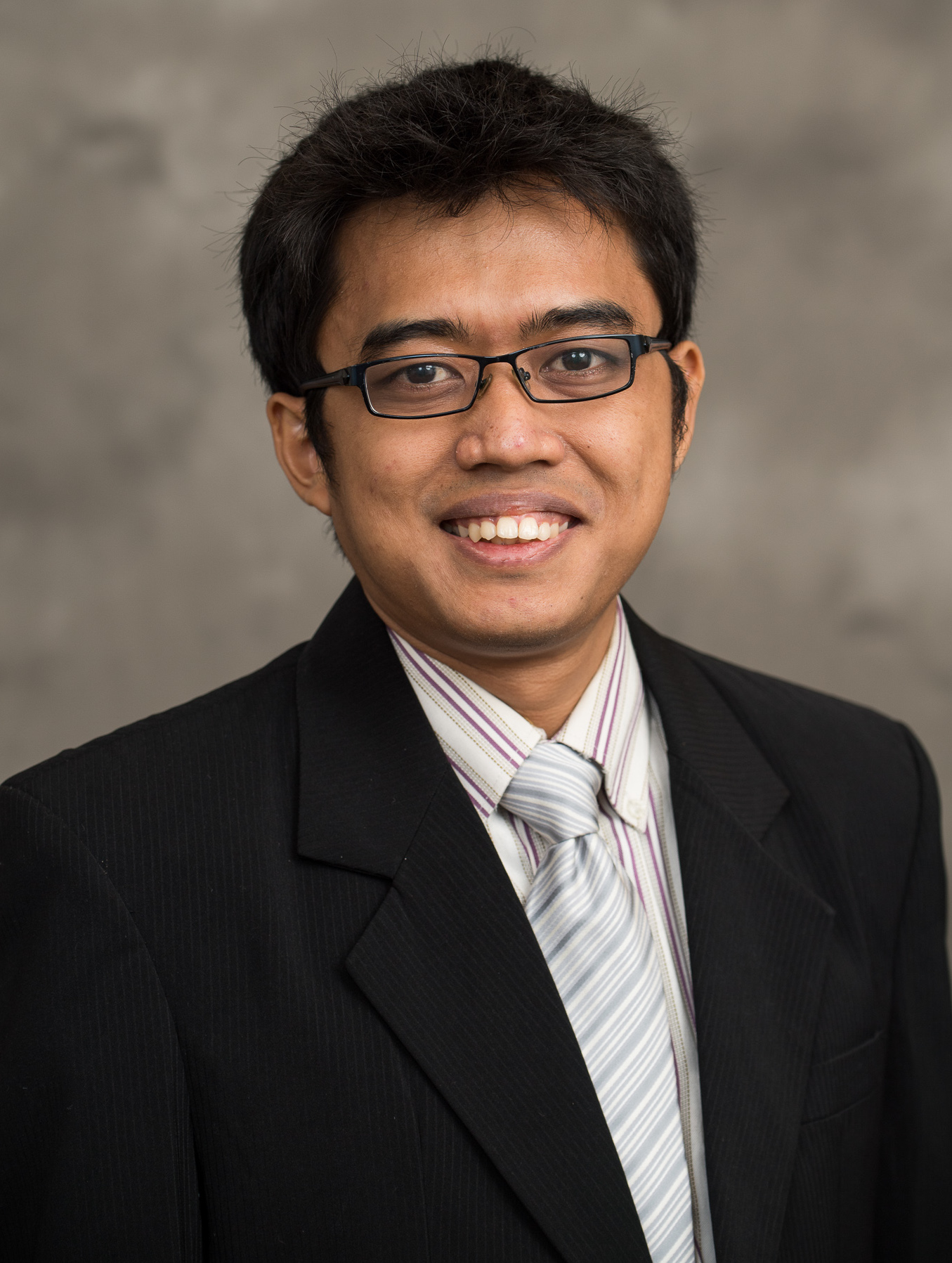}}]{Mansur Arief} is a Ph.D. student in the Safe AI Lab at Department of Mechanical Engineering, Carnegie Mellon University. His research focuses on the development of efficient rare-event simulation technique for learning-based robots using machine learning, statistical modeling, and optimization technique with applications in intelligent transportation systems.
\end{IEEEbiography}

\begin{IEEEbiography}[{\includegraphics[width=1in,height=1.25in,clip,keepaspectratio]{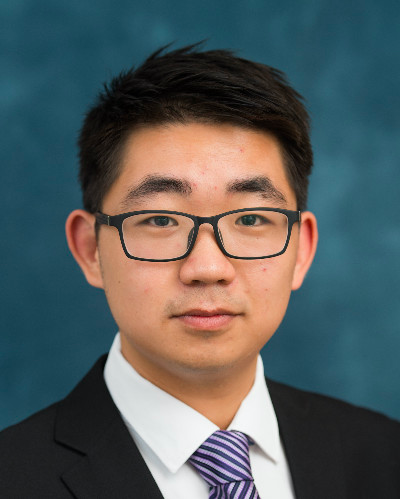}}]{Weiyang Zhang} received his B.S. degree from Zhejiang University at 2017. He now is a Master's student with the Department of Mechanical Engineering and the Department of Electrical Engineering and Computer Science, University of Michigan, Ann Arbor, MI, USA. His research focuses on autonomous driving, robotics, big data analysis, and nonparametric Bayesian learning.
\end{IEEEbiography}

\begin{IEEEbiography}[{\includegraphics[width=1in,height=1.25in,clip,keepaspectratio]{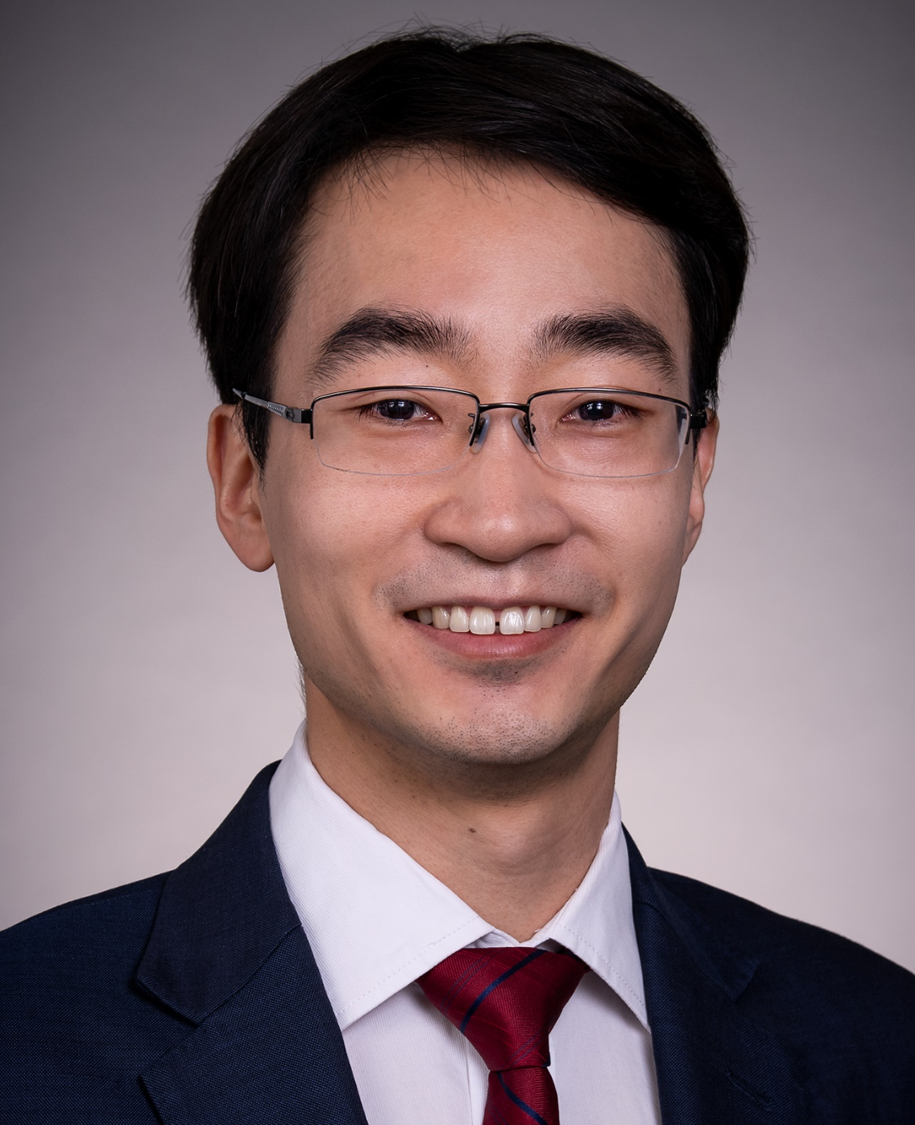}}]{Ding Zhao} received his Ph.D. degree in 2016 from
the University of Michigan, Ann Arbor. He is currently an Assistant Professor at Department of Mechanical Engineering, Carnegie Mellon University. His research aims to safely deploy the AI-enabled robots to the real world by developing reliable, verifiable, explainable, and trustworthy learning methods in the face of the uncertain, dynamic, time-varying, multi-agent, and human-involved environment. His work is at the intersection of statistical machine learning, robotics, and optimal design, with applications on autonomous vehicles, smart manufacturing, intelligent transportation, assistant robots, and cybersecurity.
\end{IEEEbiography}


\end{document}